# An Extensive Technique to Detect and Analyze Melanoma: A Challenge at the International Symposium on Biomedical Imaging (ISBI) 2017

G Wiselin Jiji[1], P Johnson Durai Raj[2]

**Abstract:** An automated method to detect and analyze the melanoma is presented to improve diagnosis which will leads to the exact treatment. Image processing techniques such as segmentation, feature descriptors and classification models are involved in this method. In the First phase the lesion region is segmented using CIELAB Color space Based Segmentation. Then feature descriptors such as shape, color and texture are extracted. Finally, in the third phase lesion region is classified as melanoma, seborrheic keratosis or nevus using multi class O-A SVM model. Experiment with ISIC 2017 Archive skin image database has been done and analyzed the results.

[1] *Professor & Principal, Dr. Sivanthi Aditanar College of Engineering, Tiruchendur-628215. Mobile: +91-9443087064, Email: jijivevin@yahoo.co.in*
[2] *Junior Research Fellow, Dr.Sivanthi Aditanar College of Engineering, Tiruchendur-628215.*

## INTRODUCTION

In the last decades the digital images produced by scientific, educational, medical, industrial and other applications are used to diagnose the various diseases. The dramatic growth of digital electronics industries has posed many challenges in dealing with huge amount of image data. The management of the expanding visual information had become a challenging task.

An effective image processing system can potentially used to classify the melanoma. The authors of [1] created a model which can characterize the pigmented skin lesion. Users can query the database by feature attribute values (shape and texture), or by synthesized image colors. It does not include a query-by example method, as do most common CBIR systems. Celebi et al. [2] developed a system for retrieving skin lesion images based on shape similarity. Rahman et al. [3] presented a CBIR system for dermatoscopic images. Their approach include image processing, segmentation, feature extraction (colour and textures) and similarity matching. Classification methods range from discriminant analysis to neural networks and support vector machines [1–4]. So based on the desired segmentation and classification of the experimenting dataset the system may opt out the image processing and machine learning techniques.

To overarching the goal of classifying the melanoma the system has three major phases, such as, Segmentation of lesion region using L*a*b color space method, extraction of feature descriptors based on Color, shape and texture and in the final phase the classification model has been developed by one-against-all SVM to classify and calculate the arbitrary score.

## CHALLENGE TASKS & DATASET

### DATA SET

The ISIC 20017 Archive Skin lesion image data based is used for experiment. In which for training phase 2000 image are used along with the ground truth. The validation data set of 150 images are used to analysis the performance. The test set of 600 image has been used to derive the final test phase submission results.

### Part 1: SEGMENTATION OF LESION REGION

The L*a*b color space based segmentation[5] is used. The L*a*b*

colorspace (also known as CIELAB or CIE L*a*b*) enable us to quantify the visual differences in lesion and normal skin region.

In our approach a small sample region is identified manually to calculate the region's average in a*b* space. This is used as the color marker to quantify the visual difference. For instance the lesion and skin region's color markers has been estimated. These two color marker now has an 'a*' and a 'b*' value.

**Nearest Neighbor Classification of Pixels**

A complete pass over the image will classify the individual pixel by calculating the Euclidean distance between that pixel and each color marker. The smallest distance will tell you that the pixel most closely matches that color marker.

If the lesion maker has the close distance then the pixel would be labeled as lesion region.

The label matrix will used to segment the lesion and skin region. Figure 1a shows the sample input image and Figure 1b shows the estimated segmentation mask for the skin lesion region by using the above method.

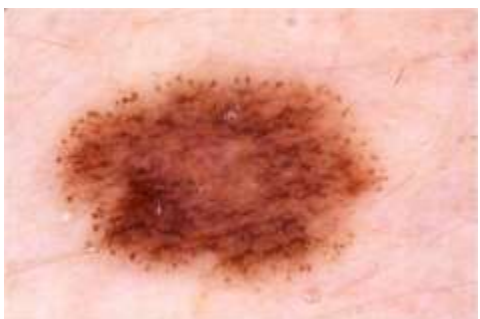

(a)

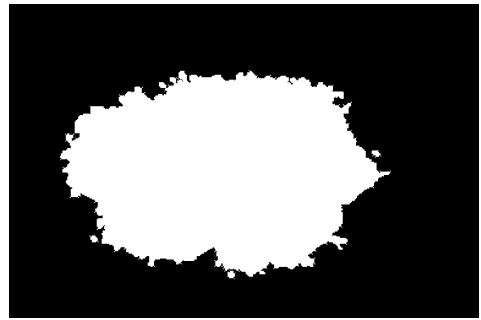

(b)

Figure 1: (a) Input Skin Image (b) Segmentation map estimated using CIELAB model.

**PART 2: FEATURE EXTRACTION**

In this section, the feature descriptions based on Color, Texture and Shape are described. All the feature descriptions were used to exemplify the skin lesion region as follows. The effectiveness of the classification is based on these visual features. To diagnosis the melanoma, seborrheic keratosis or nevus, the clinical features based on the Shape, Texture and Color are much significant and helpful.

The extraction of Shape feature and its representation are the bases to recognize the object. The shape of the object represented by their features that are used in image retrieval which consist of measuring the similarity between shapes. The shape features[6] used in this work are Area ,Perimeter, Compactness, Asymmetry, Aspect Ratio, Eccentricity, Bending Energy, Contour Moments, Invariant Moments (order 1 to 3), Convex Hull Area & Perimeter , Convexity and solidity. The textural features contains the information about the spatial distribution of tonal variations. In dermatological context, it is the major concern to distinguish the lesion type. Texture feature [6] used in this work are Fuzzy Texture Spectrum and Busyness. The property of invariant with scaling, translation and rotation is determine the feature's importance over the classification context. Color features has this property, Hence it plays the major role for classification. Color quantification and Color space are the key components of the skin image to extract the features. colour features used are Color Statistics, Mean, Standard Deviation, Moment, Co- occurrence of color indices, Contrast, Correlation, Energy, Entropy, Homogeneity.

## Part 3: CLASSIFICATION OF LESION

In Classification task, Support Vector Machine(SVM) is to produce a model which based on the training instance and can predict the target value of the test image. Given a training set of skin image's features with label pair $(x_i, y_i)$, $i=1,\ldots l$ where $x_i \in R^n$ and $y_i \in \{1,-1\}^l$, In our case the training vector $x_i$ consist of feature values. The SVM require the solution of the following optimization problem[7]:

$$\min_{w,b,\varepsilon} \frac{1}{2} w^T w + C \sum_{i=1}^{l} \varepsilon_i \qquad (1)$$

subject to $y_i(w^T \phi(x_i) + b) \geq 1 - \varepsilon_i, \quad \varepsilon_i \geq 0$ (2)

The training vectors $x_i$ are mapped into a higher dimensional space by the function $\emptyset$. SVM finds the separating hyper plane with the maximal margin in this higher dimensional space. $C > 0$ is the penalty parameter of the error term. The polynomial kernel function $K(x_i, x_j) = (1 + x_i^T x_j)^2$ is being used.

In our experiment there are two binary classification tasks. The first binary classification task belongs to melanoma vs. nevus and seborrheic keratosis. The second binary classification task belongs to Seborrheic keratosis vs. melanoma and nevus. So the One against all SVM model has been opted to attain to this said two binary classification tasks. The likelihood measure such as classification scores is represented as the arbitrary score.

## EXPERIMENTAL RESULTS

The ISIC 2017 archive has been downloaded. And the system was developed using MATLAB 2013b. The Initial segmentation task has been done for the training and validation dataset. After the submission of the validation the same thing has applied to the Test set. Feature vectors are calculated for the segmented lesion region and by make using the extracted feature vectors the SVM classification model has been trained. Finally the test set's classification results are estimated and prepared CSV file challenge submission.

The results are evaluated by the statistical performance measures such as Accuracy, Dice Coefficient, Jaccared Index, Sensitivity and Specificity. For the validation dataset the overall average performance measures are derived as 91.1% of Accuracy, 97.4% of Specificity, 66.4% sensitivity. The overall Jaccard Index score is 0.611.

## CONCLUSION

The system of automatic segmentation and classification of Melanoma is presented using the CIELAB color space and SVM classification model. The system make use the color, texture and shape features for classification task. The results shows that the system can made the segmentation task effectively based of the usual melanoma's clinical properties.